\documentclass[11pt,a4paper]{article}
\usepackage{times,latexsym}
\usepackage{url}
\usepackage[T1]{fontenc}

\usepackage[acceptedWithA]{tacl2018v2}
\usepackage{comment}
\usepackage{amsmath}
\usepackage{amsthm}
\usepackage{booktabs}
\usepackage{colortbl}
\usepackage{xspace}
\usepackage{graphicx}
\usepackage{bm}
\usepackage{subfigure}
\usepackage{bbm}
\usepackage{multirow}
\usepackage[normalem]{ulem}
\usepackage{titlesec}

\makeatletter
\renewcommand{\paragraph}{%
  \@startsection{paragraph}{4}%
  {\z@}{1.0ex \@plus 0.0ex \@minus 0.5ex}{-0.5em}%
  {\normalfont\normalsize\bfseries}%
}
\makeatother

\titlespacing*{\section}
{0pt}{1.5ex plus 0.3ex minus 0.3ex}{1.0ex plus 0.3ex minus 0.3ex}
\titlespacing*{\subsection}
{0pt}{0.7ex plus 0.3ex minus 0.3ex}{0.4ex plus 0.1ex minus 0.1ex}
\newcommand\nummark[1]{\textsuperscript#1}

\newcommand\blank{\rule[-.2ex]{7pt}{.4pt}\xspace}
\newcommand\softmax{\text{softmax}}
\newcommand\subject{x\xspace}
\newcommand\object{y\xspace}

\def\tightcol{\hskip 6pt}
\def\tinycol{\hskip 3pt}
\newcommand*{\scale}[2][4]{\scalebox{#1}{\ensuremath{#2}}}%
\newcommand{\weightsub}[1]{$_{\bm{\textcolor{blue}{#1}}}$}

\usepackage{microtype}

\title{How Can We Know What Language Models Know?}

\author{
Zhengbao Jiang\nummark{1}\thanks{~~The first two authors contributed equally.}  \quad Frank F. Xu\nummark{1}\footnotemark[1] \quad Jun Araki\nummark{2} \quad Graham Neubig\nummark{1} \\
Language Technologies Institute, Carnegie Mellon University\nummark{1} \\
Bosch Research North America\nummark{2} \\
\texttt{\{zhengbaj,fangzhex,gneubig\}@cs.cmu.edu} \quad \texttt{jun.araki@us.bosch.com}}

\date{}

\begin{document}
\maketitle
\begin{abstract}
Recent work has presented intriguing results examining the knowledge contained in language models (LM) by having the LM fill in the blanks of prompts such as ``\textit{Obama is a \blank by profession}''.
These prompts are usually manually created, and quite possibly sub-optimal; another prompt such as ``\textit{Obama worked as a \blank}'' may result in more accurately predicting the correct profession.
Because of this, given an inappropriate prompt, we might fail to retrieve facts that the LM \textit{does} know, and thus any given prompt only provides a lower bound estimate of the knowledge contained in an LM.
In this paper, we attempt to more accurately estimate the knowledge contained in LMs by automatically discovering better prompts to use in this querying process.
Specifically, we propose mining-based and paraphrasing-based methods to automatically generate high-quality and diverse prompts, as well as ensemble methods to combine answers from different prompts.
Extensive experiments on the LAMA benchmark for extracting relational knowledge from LMs demonstrate that our methods can improve accuracy from 31.1\% to 39.6\%, providing a tighter lower bound on what LMs know.
We have released the code and the resulting LM Prompt And Query Archive (LPAQA) at \url{https://github.com/jzbjyb/LPAQA}.
\end{abstract}

\section{Introduction}

\begin{figure}[t]
\center
\includegraphics[width=\columnwidth, clip, keepaspectratio]{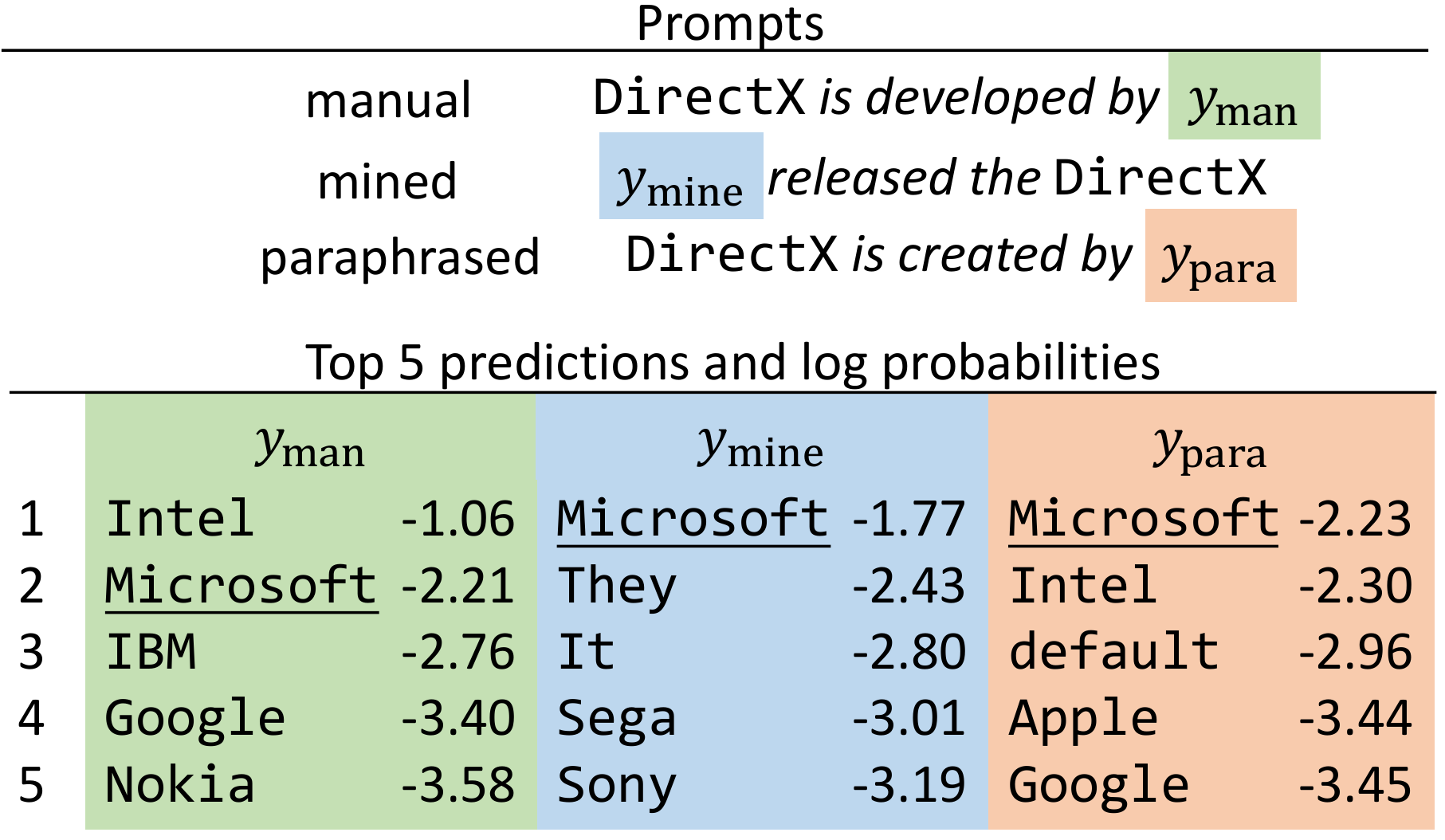}
\caption{Top-5 predictions and their log probabilities using different prompts (manual, mined, and paraphrased) to query BERT. Correct answer is underlined.}
\label{fig:example}
\end{figure}

Recent years have seen the primary role of language models (LM) transition from generating or evaluating the fluency of natural text \cite{mikolov2012context,DBLP:conf/iclr/MerityKS18,DBLP:conf/iclr/MelisDB18,gamon2005sentence} to being a powerful tool for text understanding.
This understanding has mainly been achieved through the use of language modeling as a pre-training task for \emph{feature extractors}, where the hidden vectors learned through a language modeling objective are then used in down-stream language understanding systems \cite{dai-2015-semiseq,melamud:16:context2vec,peters-2018-elmo,devlin-2019-bert}.

Interestingly, it is also becoming apparent that LMs\footnote{Some models we use in this paper, e.g.~BERT \citep{devlin-2019-bert}, are bi-directional, and do not directly define probability distribution over text, which is the underlying definition of an LM.
Nonetheless, we call them LMs for simplicity.} \emph{themselves} can be used as a tool for text understanding by formulating queries in natural language and either generating textual answers directly \cite{mccann2018natural,radford2019language}, or assessing multiple choices and picking the most likely one \cite{zweig2011microsoft,rajani-etal-2019-explain}.
For example, LMs have been used to answer factoid questions \cite{radford2019language}, answer common sense queries \cite{trinh2018simple,sap2019atomic}, or extract factual knowledge about relations between entities \cite{petroni-2019-language,baldini-soares-etal-2019-matching}.
Regardless of the end task, the knowledge contained in LMs is probed by providing a prompt, and letting the LM either generate the continuation of a prefix (e.g. ``\textit{Barack Obama was born in \blank}''), or predict missing words in a cloze-style template (e.g., ``\textit{Barack Obama is a \blank by profession}'').

However, while this paradigm has been used to achieve a number of intriguing results regarding the knowledge expressed by LMs, they usually rely on prompts that were manually created based on the intuition of the experimenter.
These manually created prompts (e.g. ``\textit{Barack Obama was born in \blank}'') might be sub-optimal because LMs might have learned target knowledge from substantially different contexts (e.g. ``\textit{The birth place of Barack Obama is Honolulu, Hawaii.}'') during their training.
Thus it is quite possible that a fact that the LM \emph{does} know cannot be retrieved due to the prompts not being effective queries for the fact.
Thus, existing results are simply a \emph{lower bound} on the extent of knowledge contained in LMs, and in fact, LMs may be even more knowledgeable than these initial results indicate.
In this paper we ask the question: ``How can we tighten this lower bound and get a more accurate estimate of the knowledge contained in state-of-the-art LMs?''
This is interesting both scientifically, as a probe of the knowledge that LMs contain, and from an engineering perspective, as it will result in higher recall when using LMs as part of a knowledge extraction system.

In particular, we focus on the setting of \citet{petroni-2019-language} who examine extracting knowledge regarding the relations between entities (definitions in \autoref{sec:lmkb}).
We propose two automatic methods to systematically improve the breadth and quality of the prompts used to query the existence of a relation (\autoref{sec:generation}).
Specifically, as shown in \autoref{fig:example}, these are \emph{mining-based} methods inspired by previous relation extraction methods \cite{ravichandran-2002-learning}, and \emph{paraphrasing-based} methods that take a seed prompt (either manually created or automatically mined), and paraphrase it into several other semantically similar expressions.
Further, because different prompts may work better when querying for different subject-object pairs, we also investigate lightweight ensemble methods to combine the answers from different prompts together (\autoref{sec:ensemble}).

We experiment on the LAMA benchmark \citep{petroni-2019-language}, which is an English-language benchmark devised to test the ability of LMs to retrieve relations between entities (\autoref{sec:experiments}).
We first demonstrate that improved prompts significantly improve accuracy on this task, with the one-best prompt extracted by our method raising accuracy from 31.1\% to 34.1\% on BERT-base \cite{devlin-2019-bert}, with similar gains being obtained with BERT-large as well.
We further demonstrate that using a diversity of prompts through ensembling further improves accuracy to 39.6\%.
We perform extensive analysis and ablations, gleaning insights both about how to best query the knowledge stored in LMs and about potential directions for incorporating knowledge into LMs themselves.
Finally, we have released the resulting LM Prompt And Query Archive (LPAQA) to facilitate future experiments on probing knowledge contained in LMs.

\section{Knowledge Retrieval from LMs}\label{sec:lmkb}
Retrieving factual knowledge from LMs is quite different from querying standard declarative knowledge bases (KB).
In standard KBs, users formulate their information needs as a structured query defined by the KB schema and query language.
For example, \texttt{SELECT ?$\object$ WHERE \{wd:Q76 wdt:P19 ?$\object$\}} is a SPARQL query to search the birth place of \texttt{Barack\_Obama}.
In contrast, LMs must be queried by natural language prompts, such as ``\textit{Barack Obama was born in \blank}'', and the word assigned the highest probability in the blank will be returned as the answer.
Unlike deterministic queries on KBs, this provides no guarantees of correctness or success.

While the idea of prompts is common to methods for extracting many varieties of knowledge from LMs, in this paper we specifically follow the formulation of \citet{petroni-2019-language}, where factual knowledge is in the form of triples $\langle \subject,r,\object \rangle$.
Here $\subject$ indicates the subject, $\object$ indicates the object, and $r$ is their corresponding relation.
To query the LM, $r$ is associated with a cloze-style prompt $t_r$ consisting of a sequence of tokens, two of which are placeholders for subjects and objects (e.g., ``\textit{$\subject$ plays at $\object$ position}'').
The existence of the fact in the LM is assessed by replacing $\subject$ with the surface form of the subject, and letting the model predict the missing object (e.g., ``\textit{LeBron James plays at \blank position}''):%
\footnote{We can also go the other way around by filling in the objects and predicting the missing subjects.
Since our focus is on improving prompts, we choose to be consistent with \citet{petroni-2019-language} to make a fair comparison, and leave exploring other settings to future work. Also notably, \citet{petroni-2019-language} only use objects consisting of a single token, so we only need to predict one word for the missing slot.}
\begin{equation*}
\hat{\object} = \arg\max_{\object^\prime \in \mathcal{V}} P_{\text{LM}}(\object^\prime|\subject, t_r),
\end{equation*}
where $\mathcal{V}$ is the vocabulary, and $P_{\text{LM}}(\object^\prime|\subject, t_r)$ is the LM probability of predicting $\object^\prime$ in the blank conditioned on the other tokens (i.e., the subject and the prompt).%
\footnote{We restrict to masked LMs in this paper because the missing slot might not be the last token in the sentence and computing this probability in traditional left-to-right LMs using Bayes' theorem is not tractable.}
We say that an LM has knowledge of a fact if $\hat{\object}$ is the same as the ground-truth $\object$.
Because we would like our prompts to most effectively elicit any knowledge contained in the LM itself, a ``good'' prompt should trigger the LM to predict the ground-truth objects as often as possible.

In previous work \citep{mccann2018natural,radford2019language,petroni-2019-language}, $t_r$ has been a single manually defined prompt based on the intuition of the experimenter.
As noted in the introduction, this method has no guarantee of being optimal, and thus we propose methods that \emph{learn} effective prompts from a small set of training data consisting of gold subject-object pairs for each relation.

\section{Prompt Generation}\label{sec:generation}

First, we tackle prompt generation: the task of generating a set of prompts $\{t_{r,i}\}_{i=1}^{T}$ for each relation $r$, where at least some of the prompts effectively trigger LMs to predict ground-truth objects.
We employ two practical methods to either mine prompt candidates from a large corpus (\autoref{sec:mine}) or diversify a seed prompt through paraphrasing (\autoref{sec:para}).

\subsection{Mining-based Generation}\label{sec:mine}

Our first method is inspired by template-based relation extraction methods \citep{agichtein-2000-snowball,ravichandran-2002-learning}, which are based on the observation that words in the vicinity of the subject $\subject$ and object $\object$ in a large corpus often describe the relation $r$.
Based on this intuition, we first identify all the Wikipedia sentences that contain both subjects and objects of a specific relation $r$ using the assumption of distant supervision, then propose two methods to extract prompts.
\paragraph{Middle-word Prompts} Following the observation that words in the middle of the subject and object are often indicative of the relation, we directly use those words as prompts.
For example, ``\textit{Barack Obama was born in Hawaii}'' is converted into a prompt ``\textit{$\subject$ was born in $\object$}'' by replacing the subject and the object with placeholders. 
\paragraph{Dependency-based Prompts}
\citet{toutanova-2015-kbtext} note that in cases of templates where words do not appear in the middle (e.g.,  ``\textit{The capital of France is Paris}''), templates based on syntactic analysis of the sentence can be more effective for relation extraction.
We follow this insight in our second strategy for prompt creation, which parses sentences with a dependency parser to identify the shortest dependency path between the subject and object, then uses the phrase spanning from the leftmost word to the rightmost word in the dependency path as a prompt.
For instance, the dependency path in the above example is ``\textit{France $\xleftarrow{\text{pobj}}$ of $\xleftarrow{\text{prep}}$ capital $\xleftarrow{\text{nsubj}}$ is $\xrightarrow{\text{attr}}$ Paris}'', where the leftmost and rightmost words are ``\textit{capital}'' and ``\textit{Paris}'', giving a prompt of ``\textit{capital of \subject is \object}''.


Notably, these mining-based methods do not rely on any manually-created prompts, and can thus be flexibly applied to any relation where we can obtain a set of subject-object pairs.
This will result in diverse prompts, covering a wide variety of ways that the relation may be expressed in text.
However, it may also be prone to noise, as many prompts acquired in this way may not be very indicative of the relation (e.g. ``\textit{\subject, \object}''), even if they are frequent.

\subsection{Paraphrasing-based Generation}\label{sec:para}

Our second method for generating prompts is more targeted -- it aims to improve lexical diversity while remaining relatively faithful to the original prompt.
Specifically, we do so by performing paraphrasing over the original prompt into other semantically similar or identical expressions.
For example, if our original prompt is ``\textit{$\subject$ shares a border with $\object$}'', it may be paraphrased into ``\textit{$\subject$ has a common border with $\object$}'' and ``\textit{$\subject$ adjoins $\object$}''.
This is conceptually similar to query expansion techniques used in information retrieval that reformulate a given query to improve retrieval performance \citep{carpineto-2012-queryexp}.

While many methods could be used for paraphrasing \cite{romano-etal-2006-investigating,bhagat-ravichandran-2008-large}, we follow the simple method of using back-translation \citep{sennrich-2016-bt,mallinson-etal-2017-paraphrasing} to first translate the initial prompt into $B$ candidates in another language, each of which is then back-translated into $B$ candidates in the original language.
We then rank $B^2$ candidates based on their round-trip probability (i.e., $P_{\text{forward}}(\bar{t}|\hat{t}) \cdot P_{\text{backward}}(t|\bar{t})$, where $\hat{t}$ is the initial prompt, $\bar{t}$ is the translated prompt in the other language, and $t$ is the final prompt), and keep the top $T$ prompts.

\section{Prompt Selection and Ensembling}\label{sec:ensemble}
In the previous section, we described methods to generate a set of candidate prompts $\{t_{r,i}\}_{i=1}^{T}$ for a particular relation $r$.
Each of these prompts may be more or less effective at eliciting knowledge from the LM, and thus it is necessary to decide how to use these generated prompts at test time.
In this section, we describe three methods to do so.

\subsection{Top-1 Prompt Selection}
For each prompt, we can measure its accuracy of predicting the ground-truth objects (on a training dataset) using:
\begin{equation*}
\scale[1.1]{
A(t_{r,i}) = \frac{\sum_{\langle \subject,\object\rangle \in \mathcal{R}}{\delta(\object=\arg\max_{\object^\prime}{P_\text{LM}(\object^\prime|\subject,t_{r,i})})}}{|\mathcal{R}|},
}
\end{equation*}
where $\mathcal{R}$ is a set of subject-object pairs with relation $r$, and $\delta(\cdot)$ is Kronecker's delta function, returning 1 if the internal condition is true and 0 otherwise.
In the simplest method for querying the LM, we choose the prompt with the highest accuracy and query using only this prompt. 

\subsection{Rank-based Ensemble}\label{sec:rank}

Next we examine methods that use not only the top-1 prompt, but combine together multiple prompts.
The advantage to this is that the LM may have observed different entity pairs in different contexts within its training data, and having a variety of prompts may allow for elicitation of knowledge that appeared in these different contexts.

Our first method for ensembling is a parameter-free method that averages the predictions of the top-ranked prompts.
We rank all the prompts based on their accuracy of predicting the objects on the training set, and use the average log probabilities%
\footnote{Intuitively, because we are combining together scores in the log space, this has the effect of penalizing objects that are very unlikely given any certain prompt in the collection.
We also compare with linear combination in ablations in \autoref{sec:loglinear}.}
from the top $K$ prompts to calculate the probability of the object:
\begin{align}
s(\object|\subject,r) &= \sum_{i=1}^{K} \frac{1}{K} \log P_{\text{LM}}(\object|\subject, t_{r,i}), \label{eq:ensemble_score} \\
P(\object|\subject, r) &= \softmax(s(\cdot|\subject,r))_{\object},
\end{align}
where $t_{r,i}$ is the prompt ranked at the $i$-th position.
Here, $K$ is a hyper-parameter, where a small $K$ focuses on the few most accurate prompts, and a large $K$ increases diversity of the prompts.

\subsection{Optimized Ensemble}\label{sec:opti}
The above method treats the top $K$ prompts equally, which is sub-optimal given some prompts are more reliable than others.
Thus, we also propose a method that directly optimizes prompt weights.
Formally, we re-define the score in \autoref{eq:ensemble_score} as:
\begin{equation}
s(\object|\subject,r) = \sum_{i=1}^{T} P_{\bm{\theta}_r}(t_{r,i}|r) \log P_{\text{LM}}(\object|\subject, t_{r,i}), \label{eq:optim_ensemble}
\end{equation}
where $P_{\bm{\theta}_r}(t_{r,i}|r) = \softmax(\bm{\theta}_r)$ is a distribution over prompts parameterized by $\bm{\theta}_r$, a $T$-sized real-value vector.
For every relation, we learn to score a different set of $T$ candidate prompts, so the total number of parameters is $T$ times the number of relations.
The parameter $\bm{\theta}_r$ is optimized to maximize the probability of the gold-standard objects $P(\object|\subject, r)$ over training data.

\section{Main Experiments}
\label{sec:experiments}

\subsection{Experimental Settings}

\begin{table}[t]
\small
\begin{center}
\begin{tabular}{c@{\tightcol}c@{\tightcol}c@{\tightcol}c}
\toprule
\textbf{Properties} & \textbf{T-REx} & \textbf{T-REx-UHN} & \textbf{T-REx-train} \\
\midrule
\#sub-obj pairs & 830.2 & 661.1 & 948.7 \\
\#unique subject & 767.8 & 600.8 & 880.1 \\
\#unique objects & 150.9 & 120.5 & 354.6 \\
object entropy & 3.6 & 3.4 & 4.4 \\
\bottomrule
\end{tabular}
\end{center}
\caption{Dataset statistics. All the values are averaged across 41 relations.}
\label{tab:data}
\end{table}

In this section, we assess the extent to which our prompts can improve fact prediction performance, raising the lower bound on the knowledge we discern is contained in LMs.

\paragraph{Dataset}
As data, we use the T-REx subset \citep{elsahar-2018-trex} of the LAMA benchmark \citep{petroni-2019-language}, which has a broader set of 41 relations (compared to the Google-RE subset which only covers 3).
Each relation is associated with at most 1000 subject-object pairs from Wikidata, and a single manually designed prompt.
To learn to mine prompts (\autoref{sec:mine}), rank prompts (\autoref{sec:rank}), or learn ensemble weights (\autoref{sec:opti}), we create a separate training set of subject-object pairs also from Wikidata for each relation that has no overlap with the T-REx dataset.
We denote the training set as T-REx-train.
For consistency with the T-REx dataset in LAMA, T-REx-train also is chosen to contain only single-token objects.
To investigate the generality of our method, we also report the performance of our methods on the Google-RE subset\footnote{\url{https://code.google.com/archive/p/relation-extraction-corpus/}}, which takes a similar form to T-REx but is relatively small and only covers 3 relations.

\citet{poerner-2019-bertnot} note that some facts in LAMA can be recalled solely based on surface forms of entities, without memorizing facts.
They filter out those easy-to-guess facts and create a more difficult benchmark, denoted as LAMA-UHN.
We also conduct experiments on the T-REx subset of LAMA-UHN (i.e., T-REx-UHN) to investigate whether our methods can still obtain improvements on this harder benchmark. Dataset statistics are summarized in \autoref{tab:data}.

\paragraph{Models}
As for the models to probe, in our main experiments we use the standard BERT-base and BERT-large models \citep{devlin-2019-bert}.
We also perform some experiments with other pre-trained models enhanced with external entity representations, i.e., ERNIE \citep{zhang-19-ernie} and KnowBert \citep{peters-2019-knowbert}, which we believe may do better on recall of entities.

\paragraph{Evaluation Metrics}
We use two metrics to evaluate the success of prompts in probing LMs.
The first evaluation metric, \emph{micro-averaged accuracy}, follows the LAMA benchmark\footnote{In LAMA, it is called ``P@1.'' There might be multiple correct answers for some cases, e.g. a person speaking multiple languages, but we only use one ground truth. We will leave exploring more advanced evaluation methods to future work.} in calculating the accuracy of all subject-object pairs for relation $r$:
\begin{equation*}
\frac{1}{|\mathcal{R}|}\sum_{\langle x,y\rangle \in \mathcal{R}}{\delta(\hat{y}=y)},
\end{equation*}
where $\hat{y}$ is the prediction and $y$ is the ground truth. Then we average across all relations.
However, we found that the object distributions of some relations are extremely skewed, e.g.~more than half of the objects in relation \texttt{native\_language} are \texttt{French}.
This can lead to deceptively high scores, even for a majority-class baseline that picks the most common object for each relation, which achieves a score of 22.0\%.
To mitigate this problem, we also report \emph{macro-averaged accuracy}, which computes accuracy for each unique object separately, then averages them together to get the relation-level accuracy:
\begin{equation*}
\frac{1}{|\text{uni\_obj}(\mathcal{R})|}\sum_{y^\prime \in \text{uni\_obj}(\mathcal{R})}{\frac{\sum_{\langle x,y\rangle \in \mathcal{R}, y=y^\prime}{\delta(\hat{y}=y)}}{|\{y|\langle x,y\rangle \in \mathcal{R}, y=y^\prime\}|}},
\end{equation*}
where $\text{uni\_obj}(\mathcal{R})$ returns a set of \emph{unique} objects from relation $r$.
This is a much stricter metric, with the majority-class baseline only achieving a score of 2.2\%.

\paragraph{Methods}
We attempted different methods for prompt generation and selection/ensembling, and compare them with the manually designed prompts used in \citet{petroni-2019-language}.
\textbf{Majority} refers to predicting the majority object for each relation, as  mentioned above.
\textbf{Man} is the baseline from \citet{petroni-2019-language} that only uses the manually designed prompts for retrieval.
\textbf{Mine} (\autoref{sec:mine}) uses the prompts mined from Wikipedia through both middle words and dependency paths, and \textbf{Mine+Man} combines them with the manual prompts.
\textbf{Mine+Para} (\autoref{sec:para}) paraphrases the highest-ranked mined prompt for each relation, while \textbf{Man+Para} uses the manual one instead.

The prompts are combined either by averaging the log probabilities from the \textbf{TopK} highest-ranked prompts (\autoref{sec:rank}) or the weights after optimization (\autoref{sec:opti}; \textbf{Opti.}).
\textbf{Oracle} represents the upper bound of the performance of the generated prompts, where a fact is judged as correct if \emph{any} one of the prompts allows the LM to successfully predict the object.
\paragraph{Implementation Details}
We use $T=40$ most frequent prompts either generated through mining or paraphrasing in all experiments, and the number of candidates in back-translation is set to $B=7$.
We remove prompts only containing stopwords/punctuations or longer than 10 words to reduce noise.
We use the round-trip English-German neural machine translation models pre-trained on WMT'19 \citep{ng-2019-fairwmt19} for back-translation, as English-German is one of the most highly resourced language pairs.%
\footnote{\url{https://github.com/pytorch/fairseq/tree/master/examples/wmt19}}
When optimizing ensemble parameters, we use Adam \citep{kingma-14-adam} with default parameters and batch size of 32.

\subsection{Evaluation Results}

\begin{table}[t]
\begin{center}
\begin{tabular}{l@{\tightcol}c@{\tightcol}c@{\tightcol}c@{\tightcol}c@{\tightcol}c}
\toprule
\textbf{Prompts} & \textbf{Top1} & \textbf{Top3} & \textbf{Top5} & \textbf{Opti.} & \textbf{Oracle} \\
\midrule
\multicolumn{6}{c}{\textit{\underline{BERT-base (\textbf{Man}=31.1)}}\vspace{1mm}} \\
\textbf{Mine} & 31.4 & 34.2 & 34.7 & 38.9 & 50.7 \\
\textbf{Mine+Man} & 31.6 & 35.9 & 35.1 & \textbf{39.6} & 52.6 \\
\textbf{Mine+Para} & 32.7 & 34.0 & 34.5 & 36.2 & 48.1 \\
\textbf{Man+Para} & \emph{34.1} & 35.8 & 36.6 & 37.3 & 47.9 \\
\midrule
\multicolumn{6}{c}{\textit{\underline{BERT-large (\textbf{Man}=32.3)}}\vspace{1mm}} \\
\textbf{Mine} & 37.0 & 37.0 & 36.4 & 43.7 & 54.4 \\
\textbf{Mine+Man} & \emph{39.4} & 40.6 & 38.4 & \textbf{43.9} & 56.1 \\
\textbf{Mine+Para} & 37.8 & 38.6 & 38.6 & 40.1 & 51.8 \\
\textbf{Man+Para} & 35.9 & 37.3 & 38.0 & 38.8 & 50.0 \\
\bottomrule
\end{tabular}
\end{center}
\caption{Micro-averaged accuracy of different methods (\%). \textbf{Majority} gives us 22.0\%. Italic indicates best single-prompt accuracy, and bold indicates the best non-oracle accuracy overall.}
\label{tab:main_micro}
\end{table}

\begin{table}[t]
\begin{center}
\begin{tabular}{l@{\tightcol}c@{\tightcol}c@{\tightcol}c@{\tightcol}c@{\tightcol}c}
\toprule
\textbf{Prompts} & \textbf{Top1} & \textbf{Top3} & \textbf{Top5} & \textbf{Opti.} & \textbf{Oracle} \\
\midrule
\multicolumn{6}{c}{\underline{\it BERT-base (\textbf{Man}=22.8)}\vspace{1mm}} \\
\textbf{Mine} & 20.7 & 22.7 & 23.9 & 25.7 & 36.2 \\
\textbf{Mine+Man} & 21.3 & 23.8 & 24.8 & \textbf{26.6} & 38.0 \\
\textbf{Mine+Para} & 21.2 & 22.4 & 23.0 & 23.6 & 34.1 \\
\textbf{Man+Para} & \emph{22.8} & 23.8 & 24.6 & 25.0 & 34.9 \\
\midrule
\multicolumn{6}{c}{\underline{\it BERT-large (\textbf{Man}=25.7)}\vspace{1mm}} \\
\textbf{Mine} & 26.4 & 26.3 & 25.9 & 30.1 & 40.7 \\
\textbf{Mine+Man} & \emph{28.1} & 28.3 & 27.3 & \textbf{30.7} & 42.2 \\
\textbf{Mine+Para} & 26.2 & 27.1 & 27.0 & 27.1 &  38.3 \\
\textbf{Man+Para} & 25.9 & 27.8 & 28.3 & 28.0 & 39.3 \\
\bottomrule
\end{tabular}
\end{center}
\caption{Macro-averaged accuracy of different methods (\%). \textbf{Majority} gives us 2.2\%. Italic indicates best single-prompt accuracy, and bold indicates the best non-oracle accuracy overall.}
\label{tab:main_macro}
\end{table}

Micro- and macro-averaged accuracy of different methods are reported in Tables \ref{tab:main_micro} and \ref{tab:main_macro} respectively.

\paragraph{Single Prompt Experiments}
When only one prompt is used (in the first \textbf{Top1} column in both tables), the best of the proposed prompt generation methods increases micro-averaged accuracy from 31.1\% to 34.1\% on BERT-base, and from 32.3\% to 39.4\% on BERT-large.
This demonstrates that the manually created prompts are a somewhat weak lower bound; there are other prompts that further improve the ability to query knowledge from LMs.

\begin{table*}[t]
\begin{center}
\begin{tabular}{llllr}
\toprule
\textbf{ID} & \textbf{Relations} & \textbf{Manual Prompts} & \textbf{Mined Prompts} & \textbf{Acc. Gain} \\
\midrule
P140 & religion & $\subject$ is affiliated with the $\object$ religion & $\subject$ who converted to $\object$ & +60.0 \\
P159 & headquarters location & The headquarter of $\subject$ is in $\object$ & $\subject$ is based in $\object$ & +4.9 \\
P20 & place of death & $\subject$ died in $\object$ & $\subject$ died at his home in $\object$ & +4.6 \\
P264 & record label & $\subject$ is represented by music label $\object$ & $\subject$ recorded for $\object$ & +17.2 \\
P279 & subclass of & $\subject$ is a subclass of $\object$ & $\subject$ is a type of $\object$ & +22.7 \\
P39 & position held & $\subject$ has the position of $\object$ & $\subject$ is elected $\object$ & +7.9 \\
\bottomrule
\end{tabular}
\end{center}
\caption{Micro-averaged accuracy gain (\%) of the mined prompts over the manual prompts.}
\label{tab:oracle_single_case}
\end{table*}

\autoref{tab:oracle_single_case} shows some of the mined prompts that resulted in a large performance gain compared to the manual ones.
For the relation \texttt{religion}, ``\textit{$\subject$ who converted to $\object$}'' improved 60.0\% over the manually defined prompt of ``\textit{$\subject$ is affiliated with the $\object$ religion}'', and for the relation \texttt{subclass\_of}, ``\textit{$\subject$ is a type of $\object$}'' raised the accuracy by 22.7\% over ``\textit{$\subject$ is a subclass of $\object$}''.
It can be seen that the largest gains from using mined prompts seem to occur in cases where the manually defined prompt is more complicated syntactically (e.g.~the former), or when it uses less common wording (e.g.~the latter) than the mined prompt.


\paragraph{Prompt Ensembling}
Next we turn to experiments that use multiple prompts to query the LM.
Comparing the single-prompt results in Column~1 to the ensembled results in the following three columns, we can see that ensembling multiple prompts almost always leads to better performance.
The simple average used in \textbf{Top3} and \textbf{Top5} outperforms \textbf{Top1} across different prompt generation methods.
The optimized ensemble further raises micro-averaged accuracy to 38.9\% and 43.7\% on BERT-base and BERT-large respectively, outperforming the rank-based ensemble by a large margin.
These two sets of results demonstrate that diverse prompts can indeed query the LM in different ways, and that the optimization-based method is able to find weights that effectively combine different prompts together.

\begin{figure}[tb]
\includegraphics[width=0.9\columnwidth, clip, keepaspectratio]{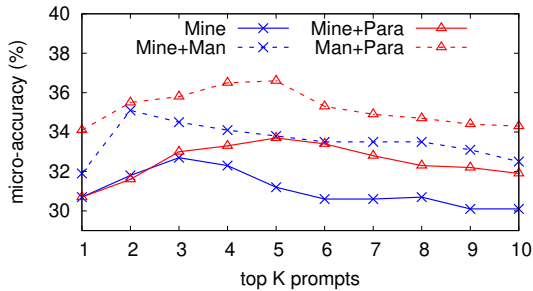}
\caption{Performance for different top-$K$ ensembles.}
\label{fig:rank}
\end{figure}

We list the learned weights of top-3 mined prompts and accuracy gain over only using the top-1 prompt in \autoref{tab:weight_case}.
Weights tend to concentrate on one particular prompt, and the other prompts serve as complements.
We also depict the performance of the rank-based ensemble method with respect to the number of prompts in \autoref{fig:rank}.
For mined prompts, top-2 or top-3 usually gives us the best results, while for paraphrased prompts, top-5 is the best.
Incorporating more prompts does not always improve accuracy, a finding consistent with the rapidly decreasing weights learned by the optimization-based method.
The gap between \textbf{Oracle} and \textbf{Opti.} indicates that there is still space for improvement using better ensemble methods.

\begin{table*}[t]
\begin{center}
\begin{tabular}{l@{\tinycol}l@{\tinycol}l@{\tinycol}r}
\toprule
\textbf{ID} & \textbf{Relations} & \textbf{Prompts and Weights} & \textbf{Acc. Gain} \\
\midrule
P127 & owned by & $\subject$ is owned by $\object$ \weightsub{.485} $\subject$ was acquired by $\object$ \weightsub{.151} $\subject$ division of $\object$ \weightsub{.151} & +7.0 \\
P140 & religion & $\subject$ who converted to $\object$ \weightsub{.615} $\object$ tirthankara $\subject$ \weightsub{.190} $\object$ dedicated to $\subject$ \weightsub{.110} & +12.2 \\
P176 & manufacturer & $\object$ introduced the $\subject$ \weightsub{.594} $\object$ announced the $\subject$ \weightsub{.286} $\subject$ attributed to the $\object$ \weightsub{.111} & +7.0 \\
\bottomrule
\end{tabular}
\end{center}
\caption{Weights of top-3 mined prompts, and the micro-averaged accuracy gain (\%) over using the top-1 prompt.}
\label{tab:weight_case}
\end{table*}

\begin{table}[tb]
\begin{center}
\begin{tabular}{llr}
\toprule
\textbf{ID} & \textbf{Modifications} & \textbf{Acc. Gain} \\
\midrule
P413 & $\subject$ plays \textcolor{blue}{in$\rightarrow$at} $\object$ position & +23.2 \\
P495 & $\subject$ was \textcolor{blue}{created$\rightarrow$made} in $\object$ & +10.8 \\
P495 & $\subject$ \textcolor{blue}{was$\rightarrow$is} created in $\object$ & +10.0 \\
P361 & $\subject$ is \textcolor{green}{a} part of $\object$ & +2.7 \\
P413 & $\subject$ plays \textcolor{red}{\sout{in}} $\object$ position & +2.2 \\
\bottomrule
\end{tabular}
\end{center}
\caption{Small modifications (\textcolor{blue}{update}, \textcolor{green}{insert}, and \textcolor{red}{delete}) in paraphrase lead to large accuracy gain (\%).}
\label{tab:para_case}
\end{table}

\paragraph{Mining vs.~Paraphrasing}
For the rank-based ensembles (\textbf{Top1, 3, 5}), prompts generated by paraphrasing usually perform better than mined prompts, while for the optimization-based ensemble (\textbf{Opti.}), mined prompts perform better.
We conjecture this is because mined prompts exhibit more variation compared to paraphrases, and proper weighting is of central importance.
This difference in the variation can be observed in the average edit distance between the prompts of each class, which is 3.27 and 2.73 for mined and paraphrased prompts respectively.
However, the improvement led by ensembling paraphrases is still significant over just using one prompt (\textbf{Top1} vs.~\textbf{Opti.}), raising micro-averaged accuracy from 32.7\% to 36.2\% on BERT-base, and from 37.8\% to 40.1\% on BERT-large.
This indicates that even small modifications to prompts can result in relatively large changes in predictions.
\autoref{tab:para_case} demonstrates cases where modification of one word (either function or content word) leads to significant accuracy improvements, indicating that large-scale LMs are still brittle to small changes in the ways they are queried.

\begin{table}[t]
\begin{center}
\begin{tabular}{l@{\tightcol}c@{\tightcol}c@{\tightcol}c@{\tightcol}c@{\tightcol}c}
\toprule
\textbf{Prompts} & \textbf{Top1} & \textbf{Top3} & \textbf{Top5} & \textbf{Opti.} & \textbf{Oracle} \\
\midrule
\textbf{Mid} & 30.7 & 32.7 & 31.2 & 36.9 & 45.1 \\
\textbf{Mid+Dep} & 31.4 & 34.2 & 34.7 & 38.9 & 50.7 \\
\bottomrule
\end{tabular}
\end{center}
\caption{Ablation study of middle-word and dependency-based prompts on BERT-base.}
\label{tab:mid_dep}
\end{table}

\paragraph{Middle-word vs. Dependency-based}
We compare the performance of only using middle-word prompts and concatenating them with dependency-based prompts in \autoref{tab:mid_dep}.
The improvements confirm our intuition that words belonging to the dependency path but not in the middle of the subject and object are also indicative of the relation.

\paragraph{Micro vs.~Macro}
Comparing \autoref{tab:main_micro} and \autoref{tab:main_macro}, we can see that macro-averaged accuracy is much lower than micro-averaged accuracy, indicating that macro-averaged accuracy is a more challenging metric that evaluates how many unique objects LMs know.
Our optimization-based method improves macro-averaged accuracy from 22.8\% to 25.7\% on BERT-base, and from 25.7\% to 30.1\% on BERT-base.
This again confirms the effectiveness of ensembling multiple prompts, but the gains are somewhat smaller.
Notably, in our optimization-based methods, the ensemble weights are optimized on each example in the training set, which is more conducive to optimizing micro-averaged accuracy.
Optimization to improve macro-averaged accuracy is potentially an interesting direction for future work that may result in prompts more generally applicable to different types of objects.

\begin{table}[t]
\begin{center}
\begin{tabular}{l@{\tightcol}c@{\tightcol}c@{\tightcol}c@{\tightcol}c@{\tightcol}c}
\toprule
\multirow{2}{*}{\textbf{Model}} & \multirow{2}{*}{\textbf{Man}} & \multirow{2}{*}{\textbf{Mine}} & \textbf{Mine} & \textbf{Mine} & \textbf{Man} \\
 &  &  & \textbf{+Man} & \textbf{+Para} & \textbf{+Para} \\
\midrule
BERT & 31.1 & 38.9 & 39.6 & 36.2 & 37.3 \\
ERNIE & 32.1 & 42.3 & 43.8 & 40.1 & 41.1 \\
KnowBert & 26.2 & 34.1 & 34.6 & 31.9 & 32.1 \\
\bottomrule
\end{tabular}
\end{center}
\caption{Micro-averaged accuracy (\%) of various LMs}
\label{tab:all_bert}
\end{table}

\begin{table}[t]
\begin{center}
\begin{tabular}{l@{\tightcol}c@{\tightcol}c@{\tightcol}c@{\tightcol}c@{\tightcol}c}
\toprule
\multirow{2}{*}{\textbf{Model}} & \multirow{2}{*}{\textbf{Man}} & \multirow{2}{*}{\textbf{Mine}} & \textbf{Mine} & \textbf{Mine} & \textbf{Man} \\
 &  &  & \textbf{+Man} & \textbf{+Para} & \textbf{+Para} \\
\midrule
BERT-base & 21.3 & 28.7 & 29.4 & 26.8 & 27.0 \\
BERT-large & 24.2 & 34.5 & 34.5 & 31.6 & 29.8 \\
\bottomrule
\end{tabular}
\end{center}
\caption{Micro-averaged accuracy (\%) on LAMA-UHN.}
\label{tab:lama_uhn}
\end{table}

\paragraph{Performance of Different LMs}
In \autoref{tab:all_bert}, we compare BERT with ERNIE and KnowBert, which are enhanced with external knowledge by explicitly incorporating entity embeddings.
ERNIE outperforms BERT by 1 point even with the manually defined prompts,  but our prompt generation methods further emphasize the difference between the two methods, with the highest accuracy numbers differing by 4.2 points using the \textbf{Mine+Man} method.
This indicates that if LMs are queried effectively, the differences between highly performant models may become more clear.
KnowBert underperforms BERT on LAMA, which is opposite to the observation made in \citet{peters-2019-knowbert}.
This is probably because that multi-token subjects/objects are used to evaluate KnowBert in \citet{peters-2019-knowbert}, while LAMA contains only single-token objects.

\paragraph{LAMA-UHN Evaluation}
The performances on LAMA-UHN benchmark are reported in \autoref{tab:lama_uhn}.
Although the overall performances drop dramatically compared to the performances on the original LAMA benchmark (\autoref{tab:main_micro}), optimized ensembles can still outperform manual prompts by a large margin, indicating that our methods are effective in retrieving knowledge that cannot be inferred based on surface forms.

\begin{table}[t]
\begin{center}
\begin{tabular}{l@{\tightcol}c@{\tightcol}c@{\tightcol}c@{\tightcol}c@{\tightcol}c}
\toprule
\multirow{2}{*}{\textbf{Model}} & \multirow{2}{*}{\textbf{Man}} & \multirow{2}{*}{\textbf{Mine}} & \textbf{Mine} & \textbf{Mine} & \textbf{Man} \\
 &  &  & \textbf{+Man} & \textbf{+Para} & \textbf{+Para} \\
\midrule
BERT-base & 9.8 & 10.0 & 10.4 & 9.6 & 10.0 \\
BERT-large & 10.5 & 10.6 & 11.3 & 10.4 & 10.7 \\
\bottomrule
\end{tabular}
\end{center}
\caption{Micro-averaged accuracy (\%) on Google-RE.}
\label{tab:googlere}
\end{table}

\paragraph{Performance on Google-RE}
We also report the performance of optimized ensemble on the Google-RE subset in \autoref{tab:googlere}.
Again, ensembling diverse prompts improves accuracies for both the BERT-base and BERT-large models.
The gains are somewhat smaller than those on the T-REx subset, which might be caused by the fact that there are only 3 relations and one of them (predicting the \texttt{birth-date} of a person) is particularly hard to the extent that only one prompt yields non-zero accuracy.

\subsection{Analysis}

\begin{figure}[t]
\center
\includegraphics[width=\columnwidth, clip, keepaspectratio]{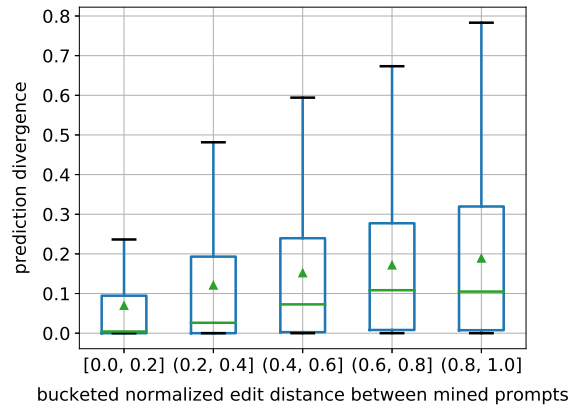}
\caption{Correlation of edit distance between prompts and their prediction divergence.}
\label{fig:editdist}
\end{figure}

Next, we perform further analysis to better understand what type of prompts proved most suitable for facilitating retrieval of knowledge from LMs.

\paragraph{Prediction Consistency by Prompt}
We first analyze the conditions under which prompts will yield different predictions.
We define the divergence between predictions of two prompts $t_{r,i}$ and $t_{r,j}$ using the following equation:
\begin{equation*}
\scale[1.0]{
\text{Div}(t_{r,i}, t_{r,j}) = \frac{\sum_{\langle \subject,\object \rangle \in \mathcal{R}}{\delta(C(\subject,\object,t_{r,i}) \ne C(\subject,\object,t_{r,j}))}}{|\mathcal{R}|},
}
\end{equation*}
where $C(\subject,\object,t_{r,i})=1$ if prompt $t_{r,i}$ can successfully predict $\object$ and $0$ otherwise, and $\delta(\cdot)$ is Kronecker's delta.
For each relation, we normalize the edit distance of two prompts into $[0,1]$ and bucket the normalized distance into 5 bins with intervals of 0.2.
We plot a box chart for each bin to visualize the distribution of prediction divergence in \autoref{fig:editdist}, with the green triangles representing mean values and the green bars in the box representing median values.
As the edit distance becomes larger, the divergence increases, which confirms our intuition that very different prompts tend to cause different prediction results.
The Pearson correlation coefficient is 0.25, which shows that there is a weak correlation between these two quantities.

\begin{table}[t]
\small
\begin{center}
\begin{tabular}{c}
\toprule
$\subject$/$\object$ V $\object$/$\subject$ \quad$|$\quad $\subject$/$\object$ V P $\object$/$\subject$ \quad$|$\quad $\subject$/$\object$ V W* P $\object$/$\subject$ \\
\midrule
V = verb particle? adv? \\
W = (noun $|$ adj $|$ adv $|$ pron $|$ det) \\
P = (prep $|$ particle $|$ inf. marker) \\
\bottomrule
\end{tabular}
\end{center}
\caption{Three part-of-speech-based regular expressions used in ReVerb to identify relational phrases.}
\label{tab:synconst}
\end{table}

\begin{figure}[t]
\center
\includegraphics[width=\columnwidth, clip, keepaspectratio]{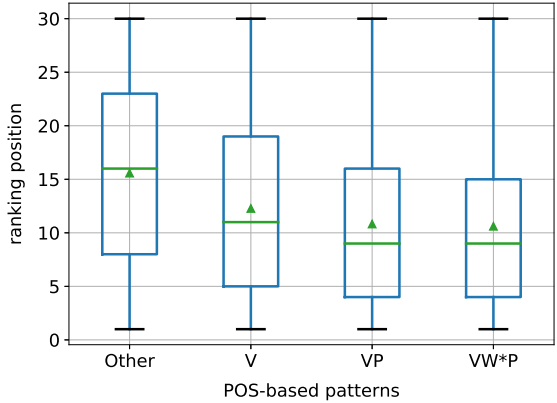}
\caption{Ranking position distribution of prompts with different patterns. Lower is better.}
\label{fig:reverb}
\end{figure}

\paragraph{POS-based Analysis}
Next, we try to examine which types of prompts tend to be effective in the abstract by examining the part-of-speech (POS) patterns of prompts that successfully extract knowledge from LMs.
In open information extraction systems \citep{banko-2007-oie}, manually defined patterns are often leveraged to filter out noisy relational phrases.
For example, ReVerb \citep{fader-2011-reverb} incorporates three syntactic constraints listed in \autoref{tab:synconst} to improve the coherence and informativeness of the mined relational phrases.
To test whether these patterns are also indicative of the ability of a prompt to retrieve knowledge from LMs, we use these three patterns to group prompts generated by our methods into four clusters, where the ``other'' cluster contains prompts that do not match any pattern.
We then calculate the rank of each prompt within the extracted prompts, and plot the distribution of rank using box plots in \autoref{fig:reverb}.%
\footnote{We use the ranking position of a prompt to represent its quality instead of its accuracy because accuracy distributions of different relations might span different ranges, making accuracy not directly comparable across relations.}
We can see that the average rank of prompts matching these patterns is better than those in the ``other'' group, confirming our intuitions that good prompts should conform with those patterns.
Some of the best performing prompts' POS signatures are ``$\subject$ VBD VBN IN $\object$'' (e.g., ``\textit{$\subject$ was born in $\object$}'') and ``$\subject$ VBZ DT NN IN $\object$'' (e.g., ``\textit{$\subject$ is the capital of $\object$}'').

\paragraph{Cross-model Consistency}
Finally, it is of interest to know whether the prompts that we are extracting are highly tailored to a specific model, or whether they can generalize across models.
To do so, we use two settings: one compares BERT-base and BERT-large, the same model architecture with different sizes; the other compares BERT-base and ERNIE, different model architectures with a comparable size.
In each setting, we compare when the optimization-based ensembles are trained on the same model, or when they are trained on one model and tested on the other.
As shown in \autoref{tab:crossmodel} and \autoref{tab:crossmodel2}, we found that in general there is usually some drop in performance in the cross-model scenario (third and fifth columns), but the losses tend to be small, and the highest performance when querying BERT-base is actually achieved by the weights optimized on BERT-large.
Notably, the best accuracies of 40.1\% and 42.2\% (\autoref{tab:crossmodel}) and 39.5\% and 40.5\% (\autoref{tab:crossmodel2}) with the weights optimized on the other model are still much higher than those obtained by the manual prompts, indicating that optimized prompts still afford large gains across models.
Another interesting observation is that the drop in performance on ERNIE (last two columns in \autoref{tab:crossmodel2}) is larger than that on BERT-large (last two columns in \autoref{tab:crossmodel}) using weights optimized on BERT-base, indicating that models sharing the same architecture benefit more from the same prompts.

\begin{table}[t]
\begin{center}
\begin{tabular}{l|cc|cc}
\toprule
\textbf{Test} & \multicolumn{2}{c|}{\textbf{BERT-base}} & \multicolumn{2}{c}{\textbf{BERT-large}} \\
\textbf{Train} & \textbf{base} & \textbf{large} & \textbf{large} & \textbf{base} \\
\midrule
Mine & 38.9 & 38.7 & 43.7 & 42.2 \\
Mine+Man & 39.6 & 40.1 & 43.9 & 42.2 \\
Mine+Para & 36.2 & 35.6 & 40.1 & 39.0 \\
Man+Para & 37.3 & 35.6 & 38.8 & 37.5 \\
\bottomrule
\end{tabular}
\end{center}
\caption{Cross-model micro-averaged accuracy (\%). The first row is the model to test, and the second row is the model on which prompt weights are learned.}
\label{tab:crossmodel}
\end{table}

\begin{table}[t]
\begin{center}
\begin{tabular}{l|c@{\tinycol}c|c@{\tinycol}c}
\toprule
\textbf{Test} & \multicolumn{2}{c|}{\textbf{BERT}} & \multicolumn{2}{c}{\textbf{ERNIE}} \\
\textbf{Train} & \textbf{BERT} & \textbf{ERNIE} & \textbf{ERNIE} & \textbf{BERT} \\
\midrule
Mine & 38.9 & 38.0 & 42.3 & 38.7 \\
Mine+Man &  39.6 & 39.5 & 43.8 & 40.5 \\
Mine+Para & 36.2 & 34.2 & 40.1 & 39.0 \\
Man+Para & 37.3 & 35.2 & 41.1 & 40.3 \\
\bottomrule
\end{tabular}
\end{center}
\caption{Cross-model micro-averaged accuracy (\%). The first row is the model to test, and the second row is the model on which prompt weights are learned.}
\label{tab:crossmodel2}
\end{table}

\begin{figure}[tb]
\includegraphics[width=0.9\columnwidth, clip, keepaspectratio]{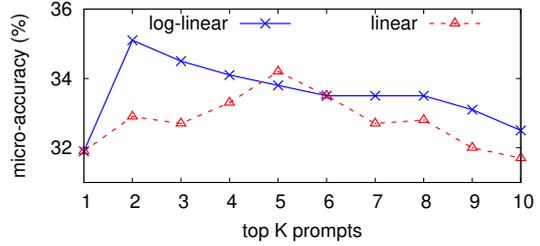}
\caption{Performance of two interpolation methods.}
\label{fig:logprob}
\end{figure}

\paragraph{Linear vs.~Log-linear Combination}\label{sec:loglinear}
As mentioned in \autoref{sec:rank}, we use log-linear combination of probabilities in our main experiments.
However, it is also possible to calculate probabilities through regular linear interpolation:
\begin{equation}
\setlength{\abovedisplayskip}{3pt}
\setlength{\belowdisplayskip}{3pt}
P(\object|\subject,r) = \sum_{i=1}^{K} \frac{1}{K} P_{\text{LM}}(\object|\subject, t_{r,i})
\end{equation}
We compare these two ways to combine predictions from multiple mined prompts in \autoref{fig:logprob} (\autoref{sec:rank}).
We assume that log-linear combination outperforms linear combination because log probabilities make it possible to penalize objects that are very unlikely given any certain prompt.

\section{Omitted Design Elements}
Finally, in addition to the elements of our main proposed methodology in \autoref{sec:generation} and \autoref{sec:ensemble}, we experimented with a few additional methods that did not prove highly effective, and thus were omitted from our final design.
We briefly describe these below, along with cursory experimental results.

\begin{table}[t]
\begin{center}
\begin{tabular}{l@{\tightcol}c@{\tightcol}c@{\tightcol}c@{\tightcol}c@{\tightcol}c}
\toprule
\textbf{Prompts} & \textbf{Top1} & \textbf{Top3} & \textbf{Top5} & \textbf{Opti.} & \textbf{Oracle} \\
\midrule
before & 31.9 & 34.5 & 33.8 & 38.1 & 47.9 \\
after & 30.2 & 32.5 & 34.7 & 37.5 & 50.8 \\
\bottomrule
\end{tabular}
\end{center}
\caption{Micro-averaged accuracy (\%) before and after LM-aware prompt fine-tuning.}
\label{tab:lmaware}
\end{table}

\subsection{LM-aware Prompt Generation}\label{sec:lmaware}
We examined methods to generate prompts by solving an optimization problem that maximizes the probability of producing the ground-truth objects with respect to the prompts:
\begin{equation*}
t_r^* = \arg\max_{t_r} P_{\text{LM}}(\object|\subject, t_r),
\end{equation*}
where $P_{\text{LM}}(\object|\subject, t_r)$ is parameterized with a pre-trained LM.
In other words, this method directly searches for a prompt that causes the LM to assign ground-truth objects the highest probability.

Solving this problem of finding text sequences that optimize some continuous objective has been studied both in the context of end-to-end sequence generation \cite{hoang-2017-towards}, and in the context of making small changes to an existing input for adversarial attacks \citep{ebrahimi-2018-hotflip,wallace-2019-universal}.
However, we found that directly optimizing prompts guided by gradients was unstable and often yielded prompts in unnatural English in our preliminary experiments.
Thus, we instead resorted to a more straightforward hill-climbing method that starts with an initial prompt, then masks out one token at a time and replaces it with the most probable token conditioned on the other tokens, inspired by the mask-predict decoding algorithm used in non-autoregressive machine translation \citep{ghazvininejad-2019-mask}:%
\footnote{In theory, this algorithm can be applied to both masked LMs like BERT and traditional left-to-right LMs, since the masked probability can be computed using Bayes' theorem for traditional LMs. However, in practice, due to the large size of vocabulary, it can only be approximated with beam search, or computed with more complicated continuous optimization algorithms \cite{hoang-2017-towards}.}
\begin{equation*}
P_{\text{LM}}(w_i|t_r\setminus i) = \frac{\sum_{\langle \subject,\object \rangle \in \mathcal{R}}{ P_{\text{LM}}(w_i|\subject,t_r\setminus i,\object)}}{|\mathcal{R}|},
\end{equation*}
where $w_i$ is the $i$-th token in the prompt and $t_r\setminus i$ is the prompt with the $i$-th token masked out.
We followed a simple rule that modifies a prompt from left to right, and this is repeated until convergence.

We used this method to refine all the mined and manual prompts on the T-REx-train dataset, and display their performance on the T-REx dataset in \autoref{tab:lmaware}.
After fine-tuning, the oracle performance increased significantly, while the ensemble performances (both rank-based and optimization-based) dropped slightly.
This indicates that LM-aware fine-tuning has the potential to discover better prompts, but some portion of the refined prompts may have over-fit to the training set upon which they were optimized.

\begin{table}[t]
\begin{center}
\begin{tabular}{l@{\tightcol}c@{\tightcol}c@{\tightcol}c@{\tightcol}c}
\toprule
\textbf{Features} & \multicolumn{2}{c}{\textbf{Mine}} & \multicolumn{2}{c}{\textbf{Paraphrase}} \\
 & macro & micro & macro & micro \\
\midrule
forward & 38.1 & 25.2 & 37.3 & 25.0 \\
\quad+backward & 38.2 & 25.5 & 37.4 & 25.2 \\
\bottomrule
\end{tabular}
\end{center}
\caption{Performance (\%) of using forward and backward features with BERT-base.}
\label{tab:backprob}
\end{table}

\subsection{Forward and Backward Probabilities}\label{sec:backprob}

Finally, given class imbalance and the propensity of the model to over-predict the majority object, we examine a method to encourage the model to predict subject-object pairs that are more strongly aligned.
Inspired by the maximum mutual information objective used in \citet{li-2016-div}, we add the backward log probability $\log P_{\text{LM}}(\subject|\object, t_{r,i})$ of each prompt to our optimization-based scoring function in \autoref{eq:optim_ensemble}.
Due to the large search space for objects, we turn to an approximation approach that only computes backward probability for the most probable $B$ objects given by the forward probability at both training and test time.
As shown in \autoref{tab:backprob}, the improvement resulting from backward probability is small, indicating that a diversity-promoting scoring function might not be necessary for knowledge retrieval from LMs.

\section{Related Work}
Much work has focused on understanding the internal representations in neural NLP models \citep{belinkov-2019-probenlpsurvey}, either by using extrinsic probing tasks to examine whether certain linguistic properties can be predicted from those representations \citep{shi-2016-string,linzen-2016-lstmsyn,belinkov-2017-neural}, or by ablations to the models to investigate how behavior varies \citep{li-2016-nnerasure,smith-2017-rnnsyn}.
For contextualized representations in particular, a broad suite of NLP tasks are used to analyze both syntactic and semantic properties, providing evidence that contextualized representations encode linguistic knowledge in different layers \citep{hewitt-2019-structprob,tenney-2019-bertpipe,tenney-2019-edgeprob,jawahar-2019-bert,goldberg-2019-assessbert}.

Different from analyses probing the representations themselves, our work follows \citet{petroni-2019-language,poerner-2019-bertnot} in probing for factual knowledge.
They use manually defined prompts, which may be under-estimating the true performance obtainable by LMs.
Concurrently to this work, \citet{bouraoui-2020-relbert} made a similar observation that using different prompts can help better extract relational knowledge from LMs, but they use models explicitly trained for relation extraction whereas our methods examine the knowledge included in LMs without any additional training.

Orthogonally, some previous works integrate external knowledge bases so that the language generation process is explicitly conditioned on symbolic knowledge \citep{ahn-2016-neulan,yang-2017-reflan,logan-2019-factlan,hayashi-2020-lrlm}.
Similar extensions have been applied to pre-trained LMs like BERT, where contextualized representations are enhanced with entity embeddings \citep{zhang-19-ernie,peters-2019-knowbert,poerner-2019-bertnot}.
In contrast, we focus on better knowledge retrieval through prompts from LMs as-is, without modifying them.

\section{Conclusion}

In this paper, we examined the importance of the prompts used in retrieving factual knowledge from language models.
We propose mining-based and paraphrasing-based methods to systematically generate diverse prompts to query specific pieces of relational knowledge.
Those prompts, when combined together, improve factual knowledge retrieval accuracy by 8\%, outperforming manually designed prompts by a large margin.
Our analysis indicates that LMs are indeed more knowledgeable than initially indicated by previous results, but they are also quite sensitive to how we query them.
This indicates potential future directions such as (1) more robust LMs that can be queried in different ways but still return similar results, (2) methods to incorporate factual knowledge in LMs, and (3) further improvements in optimizing methods to query LMs for knowledge.
Finally, we have released all our learned prompts to the community as the LM Prompt and Query Archive (LPAQA), available at: \url{https://github.com/jzbjyb/LPAQA}.

\section*{Acknowledgments}
This work was supported by a gift from Bosch Research and NSF Award No. 1815287.
We would like to thank Paul Michel, Hiroaki Hayashi, Pengcheng Yin, Shuyan Zhou for their insightful comments and suggestions.

\bibliography{tacl2020}
\bibliographystyle{acl_natbib}


\end{document}